\documentclass{article}

  \PassOptionsToPackage{numbers, compress}{natbib}
    \usepackage[preprint]{neurips_2024}

\usepackage[utf8]{inputenc} % allow utf-8 input
\usepackage[T1]{fontenc}    % use 8-bit T1 fonts
\usepackage{hyperref}       % hyperlinks
\usepackage{url}            % simple URL typesetting
\usepackage{booktabs}       % professional-quality tables
\usepackage{amsfonts}       % blackboard math symbols
\usepackage{nicefrac}       % compact symbols for 1/2, etc.
\usepackage{microtype}      % microtypography
\usepackage{xcolor}         % colors
\usepackage{graphicx}
\usepackage{multirow}
\usepackage{amsmath, bm}
\usepackage{adjustbox}
\usepackage{bbding}
\usepackage{pifont}
\usepackage{subcaption}
\usepackage{wasysym}
\usepackage{amssymb}
\usepackage{amssymb}

\title{Dual-Adapter: Training-free Dual Adaptation for Few-shot Out-of-Distribution Detection}

\author{% 
  Xinyi Chen$^{1}$ \hspace{5mm} Yaohui Li$^{1}$\hspace{5mm} Haoxing Chen$^{2}$\\
  $^{1}$Nanjing University \hspace{10mm} $^{2}$Antgroup\\
  \texttt{\{xinyichen,yaohuili\}@smail.nju.edu.cn  \hspace{5mm}} hx.chen@hotmail.com \\
}

\begin{document}

\maketitle

\begin{abstract}

We study the problem of few-shot out-of-distribution  (OOD) detection, which aims to detect OOD samples from unseen categories during inference time with only a few labeled in-domain  (ID) samples. Existing methods mainly focus on training task-aware prompts for OOD detection. However, training on few-shot data may cause severe overfitting and textual prompts alone may not be enough for effective detection. To tackle these problems, we propose a prior-based Training-free Dual Adaptation method  (\textbf{Dual-Adapter}) to detect OOD samples from both textual and visual perspectives. Specifically, Dual-Adapter first extracts the most significant channels as positive features and designates the remaining less relevant channels as negative features. Then, it constructs both a positive adapter and a negative adapter from a dual perspective, thereby better leveraging previously outlooked or interfering features in the training dataset. In this way, Dual-Adapter can inherit the advantages of CLIP not having to train, but also excels in distinguishing between ID and OOD samples. Extensive experimental results on four benchmark datasets demonstrate the superiority of Dual-Adapter.

% Few-shot OOD detection   While previous work focus on prompt learning methods, our work pioneers an approach based on model caching. Our method Tda-Adapter not only inherits the advantages of CLIP not having to train, but also excels in distinguishing between ID and OOD samples. In addition, we predict from both positive and negative perspectives to improve the performance. Extensive experiments based on ImageNet-1K demonstrates the validity and superiority of our approach.

\end{abstract}

\section{Introduction} \label{Chap: Introduction}

% 1. background  (ood detection's importance scenarios)
% 2. current methods  (prompt learning methods problem)
% 3. motivation
% 4. methods highlights advantages

In real-world settings, the deployment of machine learning models calls for the detection of out-of-distribution  (OOD) samples, as previously unseen class samples can naturally emerge and require caution.  Single-modal learning supervised methods  \cite{huang2022task, liu2020few, jeong2021few, deng2022learning, song2022few, wang2023glocal, boudiaf2023open} are widely used. However, these methods often don't make full use of text data, which can limit their effectiveness. With the advent of vision-language models like CLIP  \cite{radford2021learning},  researchers have developed numerous downstream applications using CLIP models, demonstrating significant practical value. Furthermore, these models enable a shift from traditional single-modal approaches to multi-modal learning methods, effectively addressing the OOD detection problem \cite{ming2022delving, wang2023clipn, miyai2023zero}.  

Some previous multi-modal work focus on extreme settings. Zero-shot OOD detection methods  \cite{ming2022delving, miyai2023zero} require no in-distribution (ID) training data, but may suffer from domain gaps. Fine-tuning methods  \cite{liang2017enhancing, wang2022vim, tao2023non, sun2022out}, on the other hand, require the entire training dataset, leading to high training costs. 

To address these limitations and make a trade-off between zero-shot and fine-tune methods,  some study \cite{miyai2024locoop} propose CLIP-based few-shot OOD methods, only using a few samples from training set to bridge the domain gap. Existing few-shot OOD methods primarily focus on designing task-aware prompts for OOD detection.  However, training on few-shot data can lead to overfitting, as the limited amount of data is insufficient for the model to generalize well to unseen examples. Additionally, relying solely on textual prompts may not provide enough context or information for effective OOD detection, as these prompts might not capture the full complexity of the data. More importantly, the models of these methods need to be trained, which is  time-consuming.
% 写一下之前的任务是怎么做的，有什么问题

In the field of few-shot learning, prior-based methods have been developed to address these challenges. Previous methods based on priors \cite{zhang2021tip, lin2024revisiting} can achieve high classification accuracy by leveraging CLIP priors with a cache model, eliminating the need for training.  Tip-Adapter \cite{zhang2021tip} is one of key examples of prior-based method. Unlike prompt learning, which relies heavily on textual prompts and is prone to over-fitting when trained on limited data, Tip-Adapter leverages both textual and visual features to enhance performance. However, Tip-Adapter still lacks the necessary adaptation for few-shot OOD detection, and tends to store a amount of interference information. This can lead to less effective performance, as the inference model may struggle to distinguish between relevant and irrelevant features in the limited data context.  

% Observing that, we construct positive and negative caches by distinguishing the primary features from the interfering features to address this issue. 

 % we propose a Training-free Dual Adaptation method  (\textbf{Dual-Adapter}).  Unlike prompt learning, which relies heavily on textual prompts and is prone to over-fitting when trained on limited data, Dual-Adapter leverages both textual and visual features to enhance detection performance.  Additionally, we focus on the principal subjects of the images in our training set to learn from the most relevant features. This approach is important in preventing overfitting.  

\begin{figure}[t]
  \centering
  \begin{subfigure}{0.45\linewidth} % 设定子图宽度
    \centering
    \includegraphics[width=0.99\linewidth]{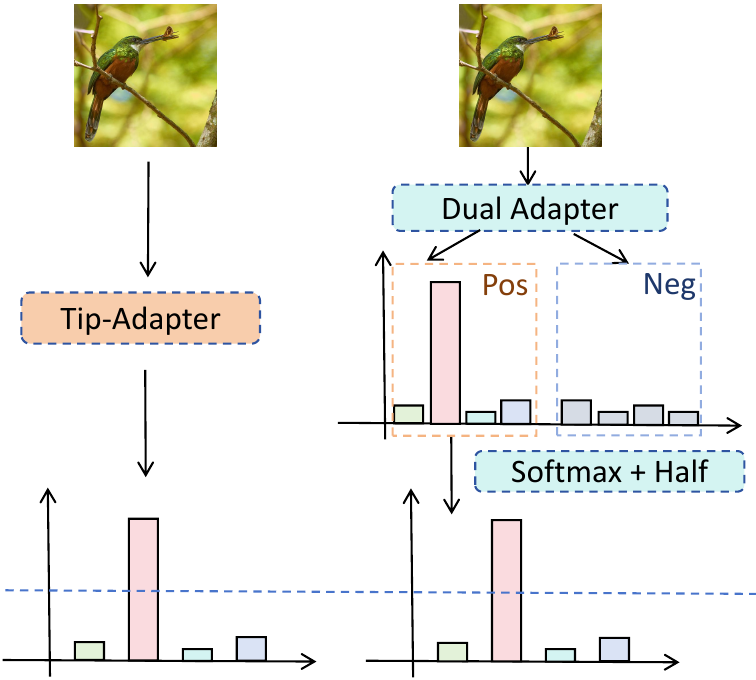} % 使用\linewidth来适应子图环境的宽度
    \caption{ID image}
    \label{figure:1-a}
  \end{subfigure}
  \begin{subfigure}{0.45\linewidth}
      \centering
    \includegraphics[width=0.99\linewidth]{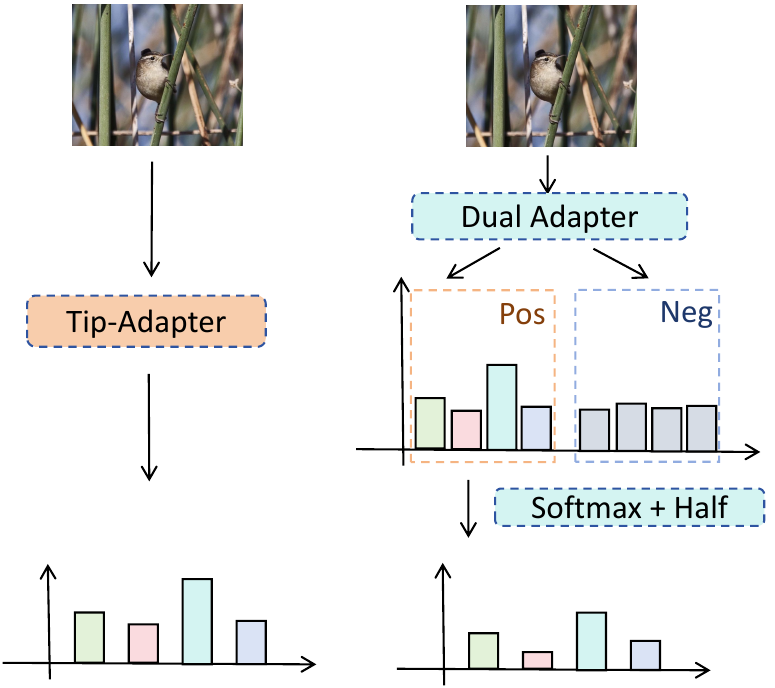} % 使用\linewidth来适应子图环境的宽度
    \caption{OOD image}
    \label{figure:1-b}
  \end{subfigure}
  \begin{subfigure}{0.45\linewidth} % 设定子图宽度
    \centering
    \includegraphics[width=0.99\linewidth]{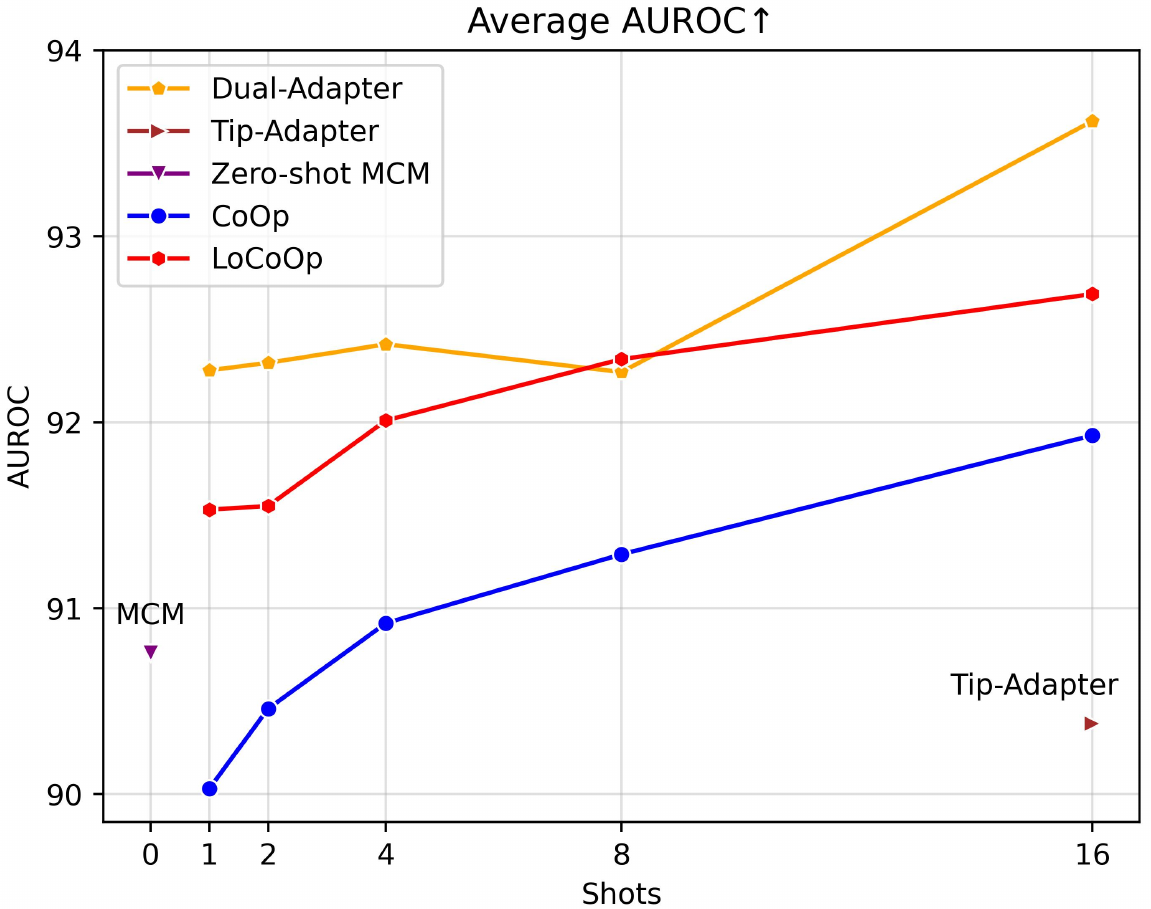} % 使用\linewidth来适应子图环境的宽度
    \caption{AUROC comparison on 4 datasets.}
    \label{figure:1-c}
  \end{subfigure}
  \begin{subfigure}{0.45\linewidth} % 设定子图宽度
    \centering
    \includegraphics[width=0.99\linewidth]{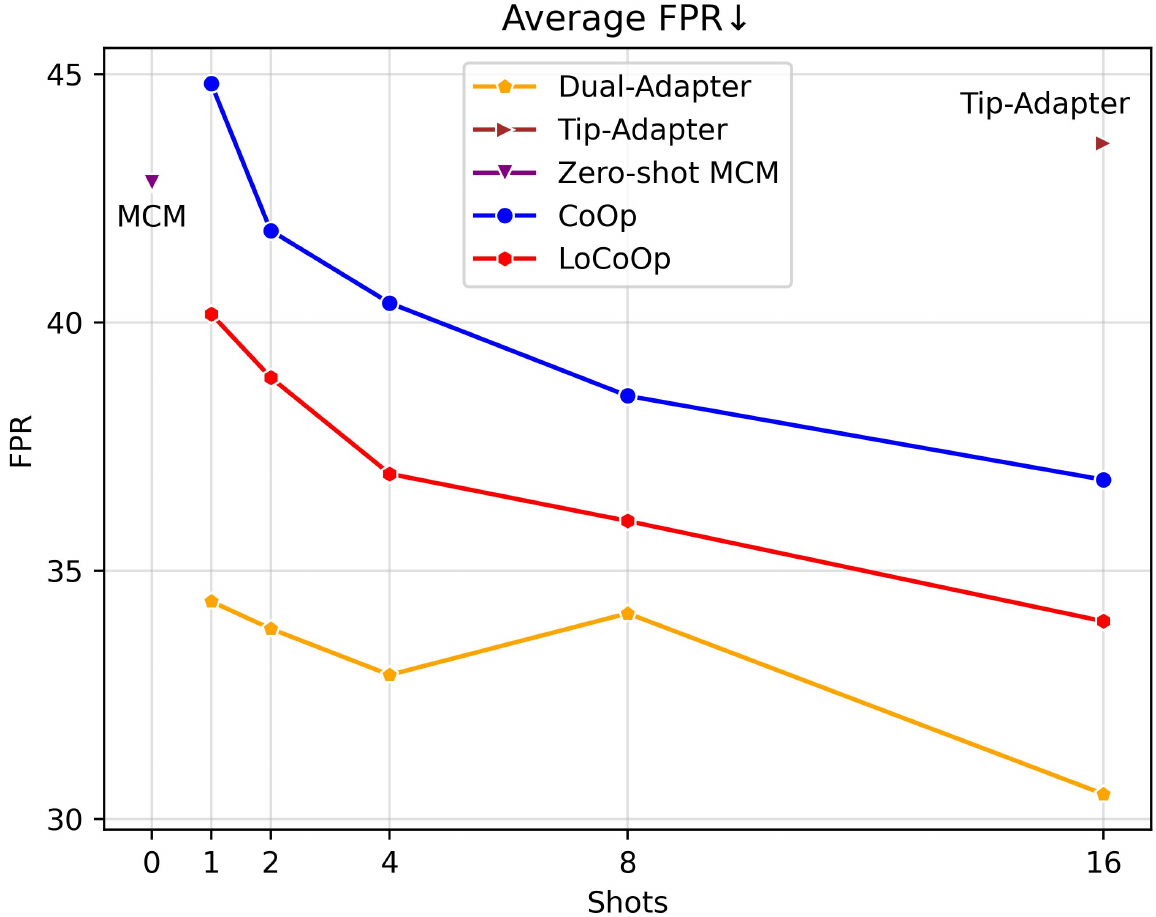} % 使用\linewidth来适应子图环境的宽度
    \caption{FPR95 comparison on 4 datasets.}
    \label{figure:1-d}
  \end{subfigure}
  \caption{\textbf{Effectiveness demonstration of Dual-Adapter.} \textbf{ (a) (b)} Compared to the performance of Tip-Adapter, Dual-Adapter remains the original logits values for ID images. In contrast, for OOD images, the logits of Dual-Adapter drop significantly. This means that Dual-Adapter can easily detect OOD samples. \textbf{ (c) (d)} Performance comparison of OOD detection, evaluated using AUROC and FPR metrics, across 1, 2, 4, 8, and 16-shot settings for Dual-Adapter, Tip-Adapter  \cite{zhang2021tip}, LoCoOp \cite{miyai2024locoop}, and MCM \cite{ming2022delving} methods.}
  \label{fig:motivation}
\end{figure}

To address the limitations of prompt learning and prior-based methods, we propose a simple yet effective approach known as the Training-free Dual Adaptation method  (\textbf{Dual-Adapter}). The motivation of our work is shown in Fig. \ref{figure:1-a} and Fig. \ref{figure:1-b}. Whereas Tip-Adapter may struggle to differentiate OOD images because of interference information, Dual-Adapter clearly distinguishes between ID and OOD images.  We employ the concept of dual cache modeling and construct Positive-Adapter and Negative-Adapter. Positive-Adapter identifies the category to which an image belongs while Negative-Adapter predicts the categories to which the image does not belong. Both adapters predict from textual and visual perspectives. Positive-Adapter and Negative-Adapters can be combined for cross-validation, enhancing performance and reliability. 

% \begin{figure}[t]
%   \centering
%   \begin{subfigure}{0.78\linewidth} % 设定子图宽度
%     \centering
%     \includegraphics[width=0.99\linewidth]{img/motivation.pdf} % 使用\linewidth来适应子图环境的宽度
%     \caption{ Comparison of Single-Adapter and Dual-Adapter.}
%     \label{figure:1-a}
%   \end{subfigure}%
%   \begin{subfigure}{0.22\linewidth} % 设定子图宽度
%     \centering
%     \includegraphics[width=0.99\linewidth]{img/avg_data.pdf} % 使用\linewidth来适应子图环境的宽度
%     \caption{Performance comparison on 4 datasets.}
%     \label{figure:1-b}
%   \end{subfigure}
%   \caption{\textbf{Effectiveness demonstration of Dual-Adapter.} \textbf{ (a)} Compared to the performance of the Single-Adapter, the Dual-Adapter remains the original logits values for ID images. In contrast, for OOD images, the logits of the Dual-Adapter drop significantly. This means that the Dual-Adapter can easily detect OOD samples. \textbf{ (b)} Comparison of OOD detection results, including AUROC and FPR metrics, across 1, 2, 4, 8, 16-shot settings among Dual-Adapter, LoCoOp \cite{miyai2024locoop}, and MCM \cite{ming2022delving} methods}
%   \label{fig:motivation}
% \end{figure}

Next, we provide an example to illustrate how Dual-Adapter effectively addresses OOD detection problems. For an image from the in-distribution  (ID) category "jacamar," the positive prediction will be significantly higher than the negative prediction, which implies it's an ID image. In contrast, for an image from an out-of-distribution  (OOD) category such as "wren", the positive prediction will be lower or approximately the same as the negative prediction, which implies it's an OOD image.

% todo：框架图overview

Dual-Adapter's performance is shown in Fig. \ref{figure:1-c} and Fig. \ref{figure:1-d}. Compared to Tip-Adapter, LoCoOp and MCM, Dual-Adapter has a great improvement in FPR and AUROC. Moreover, the most significant advantage of the Dual-Adapter is its ability to achieve optimal average performance without the need for training. This stands in stark contrast to the LoCoOp model, which requires several hours of training to reach comparable levels of effectiveness. 
% todo: 消耗少的数据支撑

% While the Vison-Language Model (VLM) has good capabilities, it lacks domain-specific adaptation, especially for domains with less available data. To perform better on downstream tasks in these areas, good use of few-shot samples is needed.
% In addition, in order to maintain the stability of the model operation and accomplish the predefined task, we need to present irrelevant interference inputs, so we are going to train a few-shot out-of-domain detector.

In summary, our contributions can be concluded as follows:
\begin{itemize}
    \item We first-time introduce prior-based methods into few-shot out-of-distribution  (OOD) detection tasks, which can identify outliers with a few labeled data without training.
    \item We introduce the concept of dual cache modeling and propose a novel methodology, Dual-Adapter, which can effectively utilize previously overlooked negative features to rectify predicted probability distributions.

    \item Extensive experimental results on benchmark datasets demonstrate the effectiveness and efficiency of Dual-Adapter.
    
\end{itemize}

\section{Related Works} \label{Chap: Related Works}
\subsection{Out-of-Distribution Detection}
Out-of-distribution  (OOD) detection is both necessary and crucial, given that deep neural networks  (DNNs) are extensively deployed in high-risk fields \cite{roy2022does, ren2019likelihood}. The presence of outliers in these areas could potentially lead to significant security incidents and substantial losses.  Previous research has extensively explored out-of-distribution  (OOD) detection in the fields of computer vision \cite{cai2023out, ge2017generative, kirichenko2020normalizing, bendale2016towards, devries2018learning} and natural language processing \cite{hu2021uncertainty, li2021k,chen2021gold, zhan2021out, jin2022towards, zhou2021contrastive}, however, these studies were mainly single-modal. Single-modal approaches are limited to utilizing either textual or visual information from samples, whereas multimodal methods can integrate both. Following the emergence of visual-language models \cite{radford2021learning, jia2021scaling, yao2021filip}, research has increasingly focused on multimodal approaches \cite{miyai2024locoop, ming2022delving, miyai2023zero, liang2017enhancing, wang2022vim, tao2023non, sun2022out,zhou2022learning}. Additionally, previous research \cite{jeong2020ood} has been constrained to small datasets due to limitations in the model's expressive capacity. By harnessing the powerful representations of  Visual-Language Models (VLM), the current approach \cite{miyai2024locoop} is better equipped to manage realistic, large-scale datasets.  Few-shot OOD detection based on VLMs offers advantages in terms of performance and application scope. 

\subsection{CLIP Prior-Based Methods}
Pre-training  Visual-Language Models (VLM) has significantly advanced multimodal learning, giving rise to numerous valuable downstream applications \cite{rao2022denseclip, jia2021scaling}. CLIP \cite{radford2021learning} is one of the most representative of these endeavours. It is popular for its simplicity and effectiveness. For its optimisation there are two categories, non-prior methods \cite{zhou2022conditional, zhou2022learning, shu2022test, gao2024clip} and prior-based methods \cite{zhang2021tip, udandarao2022sus, lin2024revisiting}. Non-prior methods add learnable modules to CLIP without using its prior knowledge, leading to limited accuracy despite introducing only a few learnable parameters. LoCoOp enables OOD detection by non-prior methods. Prior-based methods achieve superior classification performance by integrating CLIP priors with a caching model.  Tip-Adapter \cite{zhang2021tip}, one of innovative prior-based methods, acquires high-performing adapter weights without the need for training. Although previous studies have overlooked the task of out-of-distribution  (OOD) detection, our work effectively addresses this gap.

\section{Method}

\begin{figure}[t]
    \centering
    \includegraphics[width=1\linewidth]{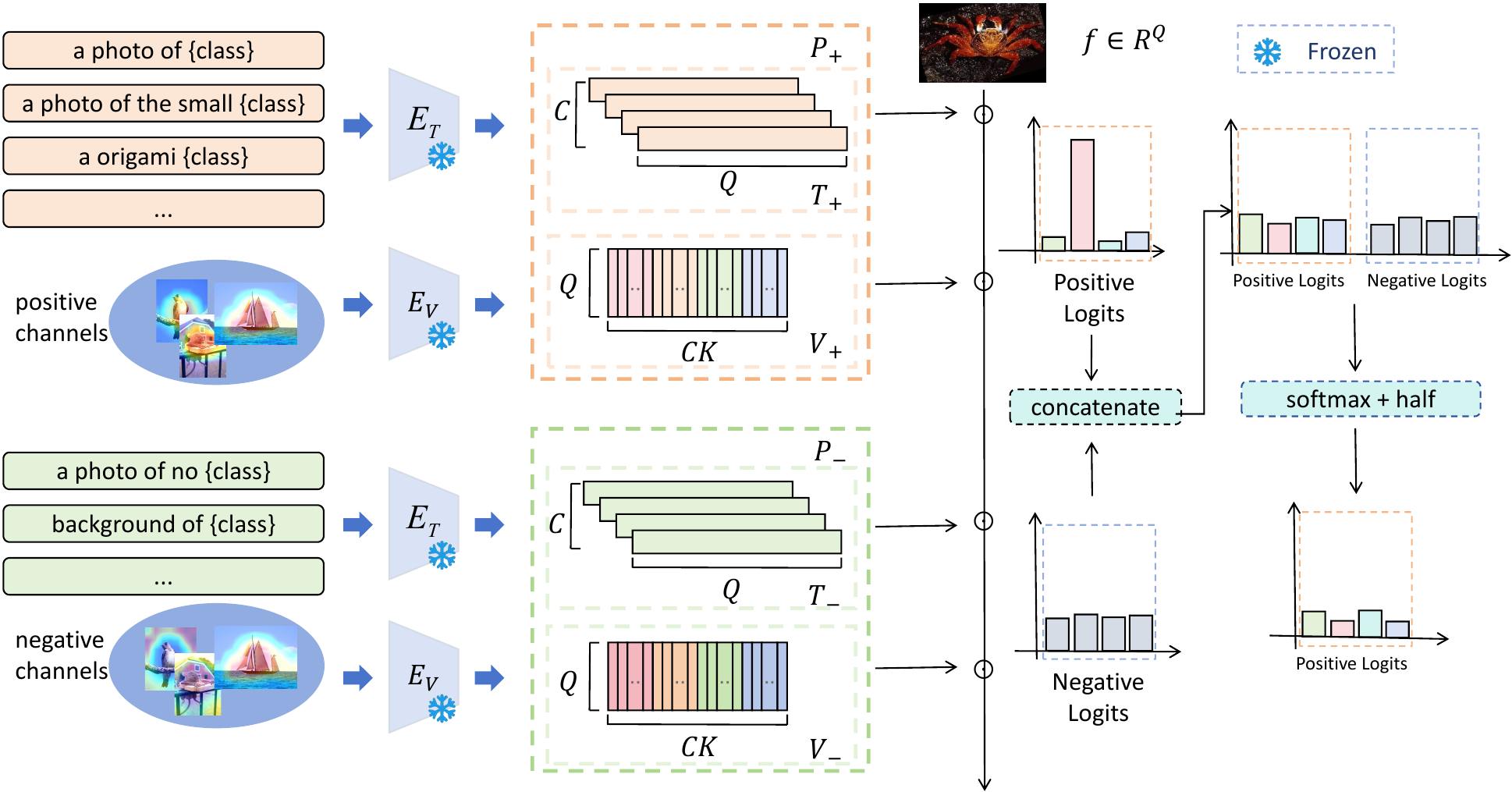}
        \caption{\textbf{An overview of Dual-Adapter.} Dual-Adapter is composed of positive adapter $P_{+}$ and negative adapter $P_{-}$, each constructed from textual and visual perspectives. These adapters further refine a new test image's features into positive and negative logits, which are concatenated and processed through a softmax function to yield the final logits for OOD detection. }
    \label{fig:overview}
\end{figure}

% TODO: f的维度注意调整
\subsection{Problem Setting}
A few-shot out-of-distribution (OOD) detector, giving a limited number of in-domain  (ID) samples, aims to distinguish between ID and OOD samples to enhance task performance in specific domains. The in-domain dataset, denoted as $D_{in}$, is used for downstream tasks and comprises a few labels training samples $x_{in}$ from specific few-shot classes $C_{in}$. The out-of-domain dataset, denoted as $D_{out}$, contains task-irrelevant samples $x_{out}$  from OOD classes $C_{out}$. Formally, $C_{in} \cap C_{out} = \emptyset$.

To build the cache model, we randomly selected 1, 2, 4, 8 or 16 samples per class from $D_{in}$ for different few-shot settings. Our training-free Dual-Adapter, relies solely on pretrained model and the cache model for prediction without any access to $D_{out}$. During testing, $D_{in}$ and $D_{out}$ are combined to evaluate the model's performance. 

\subsection{Reviews of Tip-Adapter}

Tip-adapter is an innovative CLIP-based approach that achieves superior classification performance by constructing cached key-value models from only few-shot samples of the training set without backpropagating the training adapter.  Tip-adapter utilizes the CLIP image encoder, $E_{V}: \mathbb{R}^{H \times W \times 3} \rightarrow \mathbb{R}^{N}$, and the text encoder, $E_{T}: \mathbb{R}^{m \times N_{T}} \rightarrow \mathbb{R}^{N}$, to transform visual features and textual features into a unified $N$-dimensional embedding space.  

For a $C$-category dataset, Tip-adapter selects $K$ images per class and uses the CLIP image encoder $E_{V}$ to extract $N$-dimensional features from each image. The key-value cache model is initialized using these image features, denoted as $V \in \mathbb{R}^{CK \times N}$, and corresponding one-hot label, denoted as $L \in \mathbb{R}^{CK \times C}$.   Additionally, the CLIP text encoder $E_{T}$ is used to extract the textual features of the categories $C$ to construct the zero-shot classifier $T \in \mathbb{R}^{C \times N}$, which includes the textual features of each category. 

For a test image sample from $D_{in}$ or $D_{out}$, we extract the image's visual features $f \in \mathbb{R}^{N}$ . The zero-shot prediction is calculated as
\begin{equation}
    S_{fT} = fT^T \in \mathbb{R}^{1 \times C}.
    \label{eq:zero}
\end{equation}
The similarity between test image and the cache model is calculated as
\begin{equation}
    S_{fV} = \exp (- \beta  (1 - fV^T))\in \mathbb{R}^{1 \times CK} ,
    \label{eq:cache}
\end{equation}
where $\beta$  serves as a regulatory hyperparameters. The final few-shot prediction combines the zero-shot prediction and the cache-based prediction: 
\begin{equation}
    logits = S_{fT} + \alpha S_{fV} L \in \mathbb{R}^{1 \times C},
    \label{eq:logits}
\end{equation}
where $\alpha$ serves as a balance parameter to regulate the weight of zero-shot prediction and cache-based prediction.  Subsequently, the ultimate prediction is derived by applying softmax  on logits.

\subsection{Proposed Approach}
In this section, we introduce our innovative training-free \textbf{Dual-Adapter} approach to solve few-shot out-of-distribution (OOD) problems, as illustrated in Fig. \ref{fig:overview}. The Dual-Adapter consists of two components: the Positive-Adapter and the Negative-Adapter, which are built from selected training set features and work together to make predictions.  

\noindent
\paragraph{Feature extraction} Both the CLIP image encoder $E_{V}$ and text encoder $E_{T}$ map inputs into an $N$-dimensional embedding, with each embedding containing $N$ channels . However, not all  channels really matters and are relevant to downstream tasks. We calculate inter-class similarity $S_{i}$ and inter-class variance $V_{i}$ for the $i$-th channel, wanting to minimize inter-class similarity and maximize inter-class variance. The feature importance indicator is calculated as
\begin{equation}
    F_{i} = \lambda S_{i} +  (1 - \lambda) V_{i},
    \label{eq:feature_importance}
\end{equation}
where $\lambda$ is a balance factor and $i = 1,2,...,N$. The smaller $F_{i}$ is, the more influential the $i$-th channel is. We select the most influential $Q (Q = N / 2)$ channels as positive channels $C_{pos}$, while the remaining channels are used as negative channels $C_{neg}$. The positive channels widen the distance between sample representations. To match the dimensions, we perform a pooling operation on the image visual features $f$ to extract $Q$-dimensional features. The extracted image features, denoted as $f_{Q}$, replaces $f$ in the subsequent equations.

\noindent
\paragraph{Positive-Adapter}
For the Positive-Adapter, the positive cache model is built from the positive channels $C_{pos} \in \mathbb{R}^{Q}$. Given a C-category K-shot training dataset, the positive cache model stores positive features $V_{+} \in \mathbb{R}^{CK \times Q}$. The corresponding text features are extracted from the positive template "a class of {category}" to build positive zero-shot classifier $T_{+}$.  Replace $T$ in Eq. \ref{eq:zero} with $T{+}$to calculate the positive zero-shot prediction, and replace $V$ in Eq. \ref{eq:cache} with $V_{+}$ to calculate the positive cached prediction:
\begin{equation}
    S_{fT+} = f_{Q}T_{+}^T \in \mathbb{R}^{1 \times C},
    \label{eq:pos-zero}
\end{equation}
\begin{equation}
    S_{fV+} = \exp (- \beta  (1 - f_{Q}V_{+}^T))\in \mathbb{R}^{1 \times CK}.
    \label{eq:pos-cache}
\end{equation}
The formula for calculating the prediction of each category $C_{in}$  from the in-domain dataset $D_{in}$ is:
\begin{equation}
    P_{+} = S_{fT+} + \alpha S_{fV+} L \in \mathbb{R}^{1 \times C}.
    \label{eq:pos-logits}
\end{equation}

% \begin{figure}
%     \centering
%     \includegraphics[width=0.9\linewidth]{img/channels_negative.pdf}
%     \caption{\textbf{Heatmap for extracted negative channels}}
%     \label{fig:channels_negative}
% \end{figure}

\noindent
\paragraph{Negative Adapter}
For Negative-Adapter, the negative cache model is built from the negative channels $C_{neg} \in \mathbb{R}^{Q}$. Given a C-category K-shot training dataset, the negative cache model stores negative features $V_{-} \in \mathbb{R}^{CK \times Q}$. The corresponding text features are extracted from the negative template "a class of no {category}" or "background of {category}" to build negative zero-shot classifier $T_{-}$.  The exact negative template to use is determined by the OOD dataset. Replace $T$ in Eq. \ref{eq:zero} with $T_{-}$to calculate negative zero-shot prediction, and replace $V$ in Eq. \ref{eq:cache} with $V_{-}$ to calculate negative cached prediction:
\begin{equation}
    S_{fT-} = f_{Q}T_{-}^T \in \mathbb{R}^{1 \times C},
    \label{eq:neg-zero}
\end{equation}
\begin{equation}
    S_{fV-} = \exp (- \beta  (1 - f_{Q}V_{-}^T))\in \mathbb{R}^{1 \times CK}.
    \label{eq:neg-cache}
\end{equation}
The formula for calculating the prediction of not belonging to each category $C_{in}$  from in-domain dataset $D_{in}$ is:
\begin{equation}
    P_{-} = S_{fT-} + \alpha S_{fV-} L \in \mathbb{R}^{1 \times C}.
    \label{eq:pos-logits}
\end{equation}

\noindent
\paragraph{Dual-Adapter}
The Positive-Adapter yields the probability that the sample belongs to ID categories $C_{in}$, in contrast, the Negative-Adapter yields the probability that the sample doesn't belong to ID categories $C_{in}$. Dual-Adapter combines results from Positive-Adapter and Negative-Adapter to enhance OOD detection.

In the previous selection of feature channels, the positive channels would distance the samples.  So for a ID sample from $D_{in}$, positive adapters will be more sensitive to its characteristics, the distribution of logits would be steeper. Therefore $P_{+}$ should be generally higher than $P_{-}$.  In contrast, for a sample from $D_{out}$, which is totally unseen from Dual-Adapter,  both positive and negative adapters are sluggish to its features, the distribution of logits would be relatively flat. Consequently, $P_{+}$ and $P_{-}$ would be closer.

We concatenate $P_{+}$ and $P_{-}$ as $P_{dual}$, then apply softmax function:
\begin{equation}
    P_{dual} = softmax ([P_{+}, P_{-}], \tau) \in \mathbb{R}^{1 \times 2C},
    \label{eq:con-softmax}
\end{equation}
where $\tau$ is temperature parameter to control the sharpness of the output probability distribution. 
The default value of $\tau$ is set to 1 here  A higher temperature leads to a flatter, more uniform distribution, while a lower temperature results in a sharper distribution, emphasizing the highest values.  
As a result of concatenation and softmax function, negative prediction $P_{-}$ diverts attention from positive prediction $P_{+}$, aiding in subsequent OOD detection.  As Fig. \ref{figure:1-a} and Fig. \ref{figure:1-b} shown, the logits distribution for an in-distribution  (ID) image remains largely unchanged, whereas for an out-of-distribution  (OOD) image, the logits distribution sharply decreases. This mechanism enhances the model's ability to distinguish between ID and OOD samples. 

\subsection{Test-time OOD Detection}
In testing, we adopt MCM \cite{ming2022delving} (maximum concept matching) method to evaluate. For an input image $x$, MCM calculate the similarity between the image feature $f_x \in \mathbb{R}^{N}$ and each category' textual features $t_{i} \in \mathbb{R}^{N}$. Previous zero-shot classifier $T\in \mathbb{R}^{C \times N}$ is concatenation of all $t_{i}$ .Then take the maximum similarity.  Formally, 
\begin{equation}
    \mathcal{L}_{\text{MCM}} = \max_i\frac{\exp (\text{sim} (f_x, t_i)/\tau)}{\sum_{i'=1}^C \exp (\text{sim} (f_x, t_{i'})/\tau)} .
    \label{eq:mcm}
\end{equation}

\section{Experiment} \label{Chap: Experiment}
In this section, we perform extensive experiments on out-of-distribution detection over four different dataset. In Section  \ref{Experimental Setting}, we provide the implementation details of the experiments, including information on the datasets, devices, backbones, hyper-parameters, comparison methods, and evaluation metrics.  In Section  \ref{Results}, we analyze the experimental results across different datasets and various few-shot settings.

\subsection{Experimental Setting}  \label{Experimental Setting}
\paragraph{Dataset} 
We conduct experiments for Dual-Adapter using Imagenet-1K \cite{deng2009imagenet} as the in-domain dataset, subset of iNatualist \cite{van2018inaturalist}, SUN \cite{xiao2010sun}, Places \cite{zhou2017places}, Texture \cite{cimpoi2014describing} as the out-of-domain datasets. These datasets represent a wider range of scenarios.  For few-shot training, we use the evaluation protocol adopted in CLIP  \cite{radford2021learning} , CoOp  \cite{zhou2022learning} and LoCoOp \cite{miyai2024locoop}, training with 1, 2, 4, 8, and 16 shots. Models are then evaluated on the full test sets.  For comparison, we report the average results across three runs. 

\paragraph{Setup} \label{Chap: Setup}
Following previous work  \cite{tao2023non, miyai2024locoop}, we employ ViT-B/16 \cite{deng2009imagenet} as as the backbone for both the text encoder and the image encoder.  This model extracts 512-channel features from category text information and images. We partition these features into two sets: 256-channel features are designated as positive features, while the remaining 256-channel features are considered negative features.  For positive text prompts in Imagenet-1K \cite{deng2009imagenet} , we utilize the textual prompts from Tip-X \cite{udandarao2022sus} and CuPL \cite{pratt2023does}. To generate negative text prompts, we input templates "a photo of {class}" and "background of {class}" into CLIP textual encoder, then we take the average to represent the negative category.  All experiments are conducted using a single Tesla V100-SXM2-32GB.  We vary the parameter $\tau$ from 0.03 to 10, and adjust $\alpha$ and $\beta$ from 0.0 to 30.0. During inference, we search for the optimal values of $\tau$, $\alpha$ and $\beta$.

\paragraph{Baseline methods} 
To demonstrate the superiority of our approach, we conduct a comparative analysis with the pioneering CLIP-based few-shot out-of-distribution  (OOD) detection method, CoOp  \cite{zhou2022learning}  and LoCoOp \cite{miyai2024locoop}.   LoCoOp has been shown to outperform zero-shot detection methods  \cite{ming2022delving, miyai2023zero}, fully-supervised detection methods  \cite{liang2017enhancing, wang2022vim, tao2023non, sun2022out}, and the baseline prompt learning method \cite{zhou2022learning}. By comparing our method against LoCoOp, we aim to highlight the advancements and enhancements our approach offers in the realm of OOD detection. 

\paragraph{Evaluation Metrics}
To measure the OOD performance, we adopt the following metics:   

 (1) \textbf{FPR95} (false positive rate at 95\% true positive rate): This metric measures the false positive rate  (FPR) of OOD images when the true positive rate  (TPR) of ID images is set to 95\%.  

 (2) \textbf{AUROC} (the area under the receiver operating characteristic curve): This metric evaluates the model's ability to distinguish between positive and negative classes. The ROC curve plots the true positive rate  (TPR) against the false positive rate  (FPR) at various thresholds. AUROC ranges from 0 to 1, with 1 indicating perfect discrimination and 0.5 indicating no discrimination  (random guessing).

\begin{figure}[t]
    \centering
    \begin{subfigure}{0.245\linewidth}
        \captionsetup{aboveskip=0pt, belowskip=5pt}
        \includegraphics[width=0.99\linewidth]{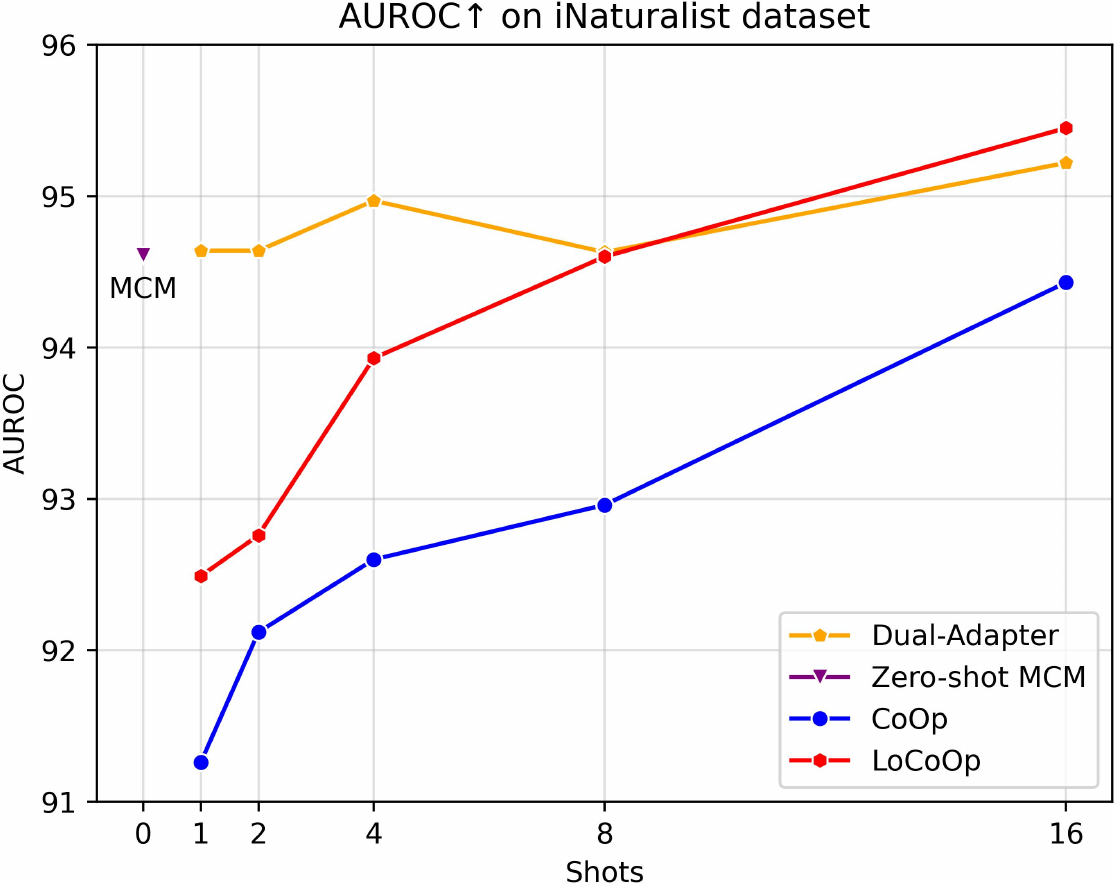}
        \label{fig:results-1}
        \caption{iNaturalist AUROC}
    \end{subfigure}
    \begin{subfigure}{0.245\linewidth}
        \captionsetup{aboveskip=0pt, belowskip=5pt}
        \includegraphics[width=0.99\linewidth]{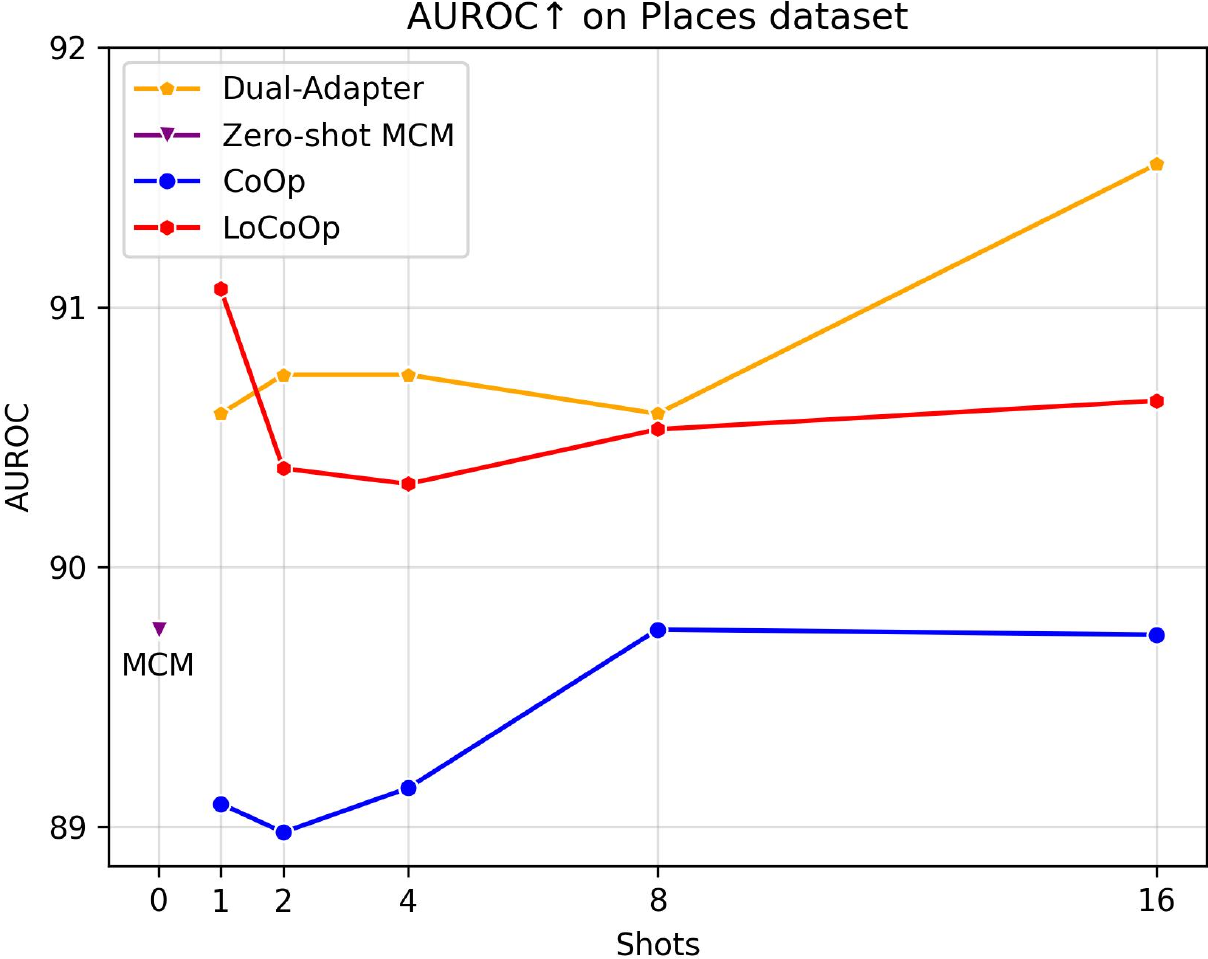}
        \label{fig:results-2}
        \caption{Places AUROC}
    \end{subfigure}
    \begin{subfigure}{0.245\linewidth}
        \captionsetup{aboveskip=0pt, belowskip=5pt}
        \includegraphics[width=0.99\linewidth]{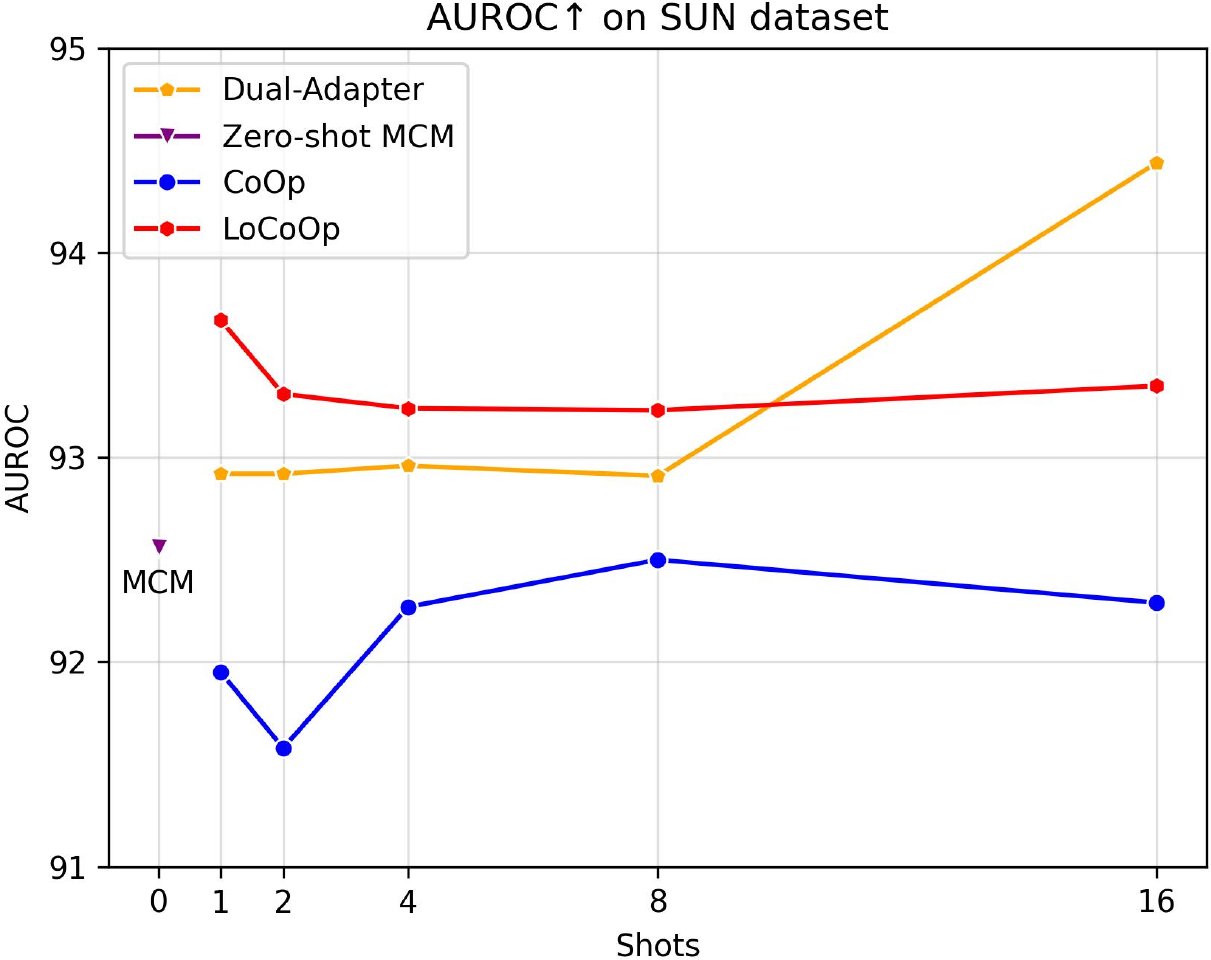}
        \label{fig:results-3}
        \caption{SUN AUROC}
    \end{subfigure}
    \begin{subfigure}{0.245\linewidth}
        \captionsetup{aboveskip=-2pt, belowskip=5pt}
        \includegraphics[width=0.99\linewidth]{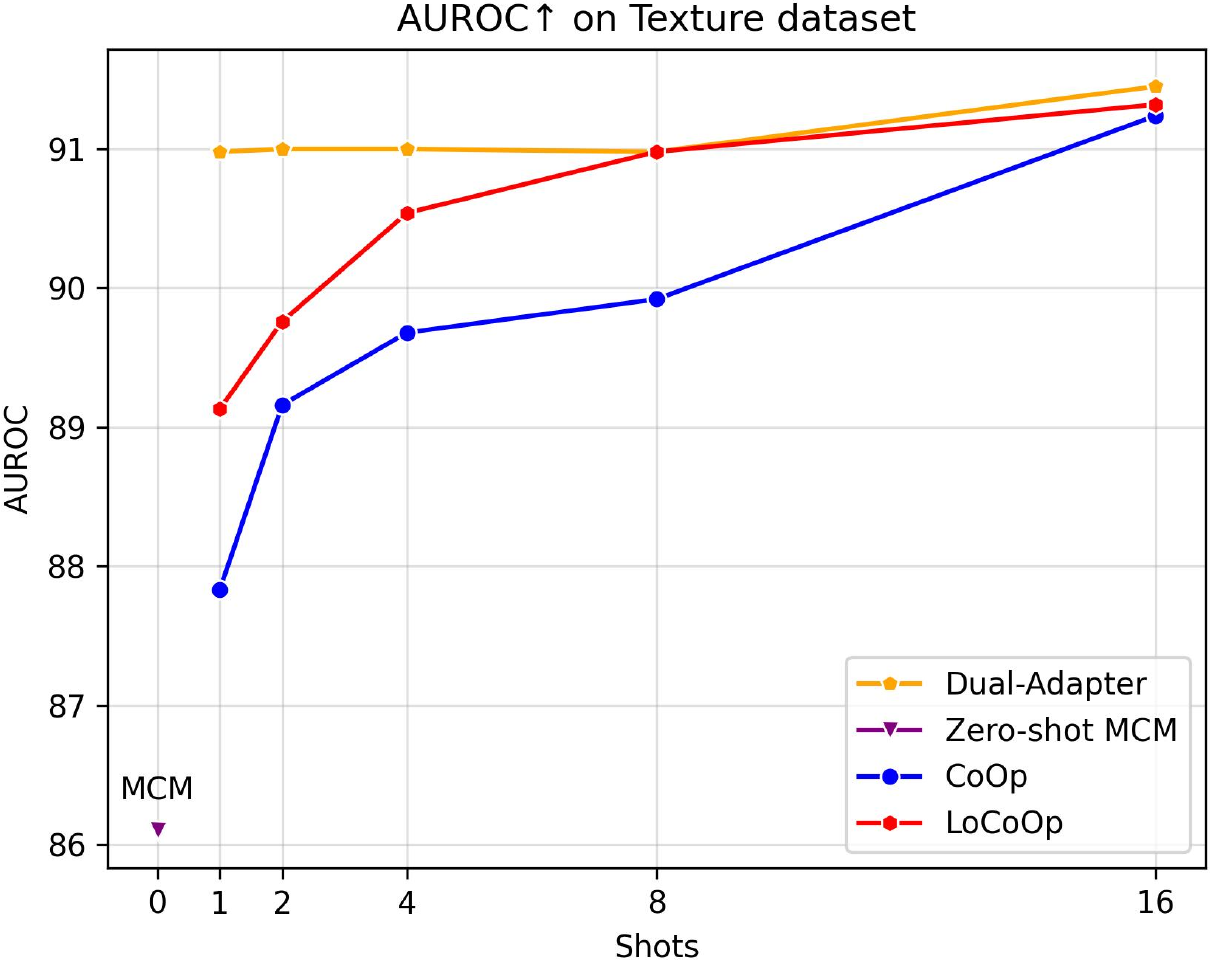}
        \label{fig:results-4}
        \caption{Texture AUROC}
    \end{subfigure}
    \begin{subfigure}{0.245\linewidth}
        \captionsetup{aboveskip=-2pt, belowskip=0pt}
        \includegraphics[width=0.99\linewidth]{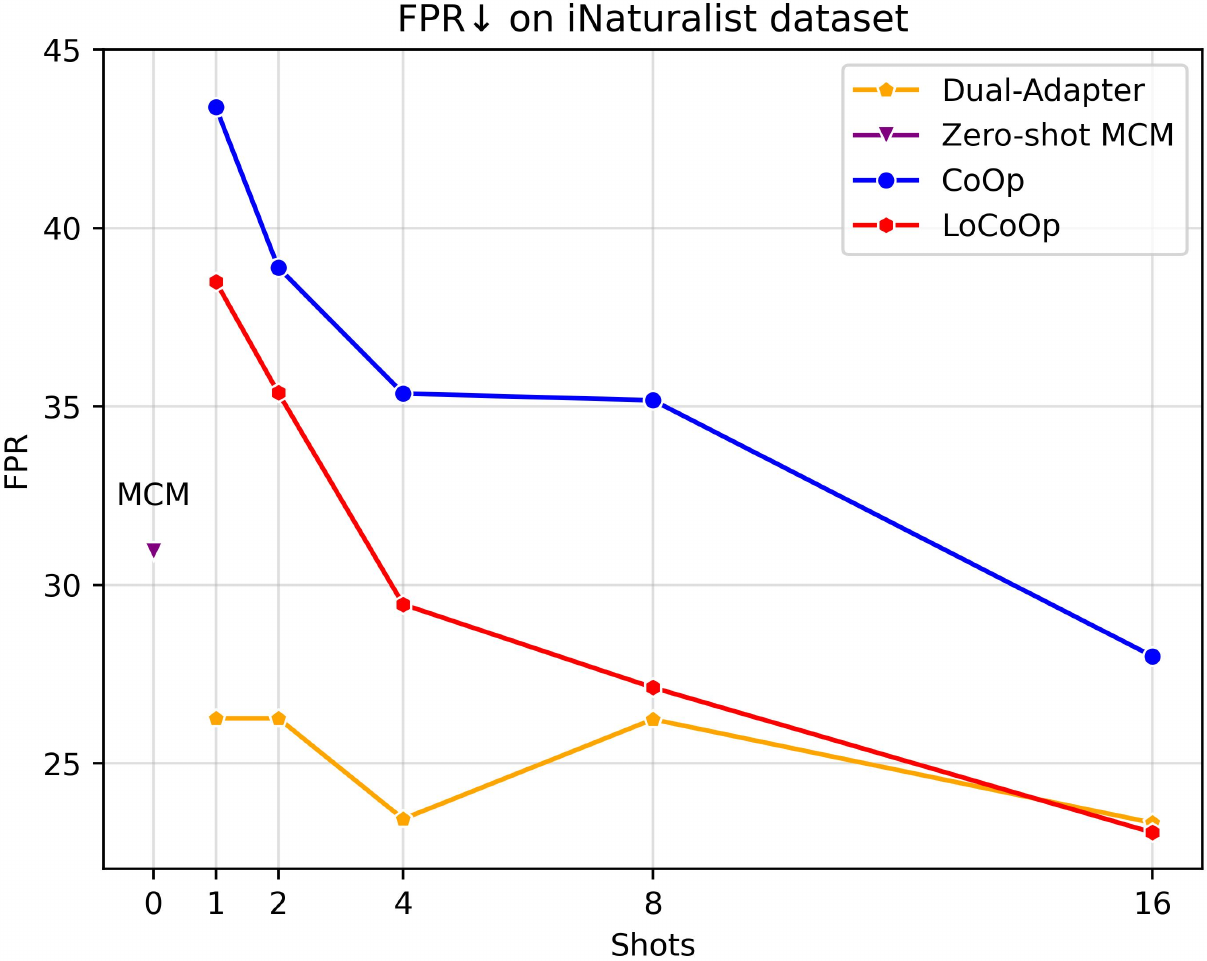}
        \label{fig:results-5}
        \caption{iNaturalist FPR}
    \end{subfigure}
    \begin{subfigure}{0.245\linewidth}
        \captionsetup{aboveskip=-2pt, belowskip=0pt}
        \includegraphics[width=0.99\linewidth]{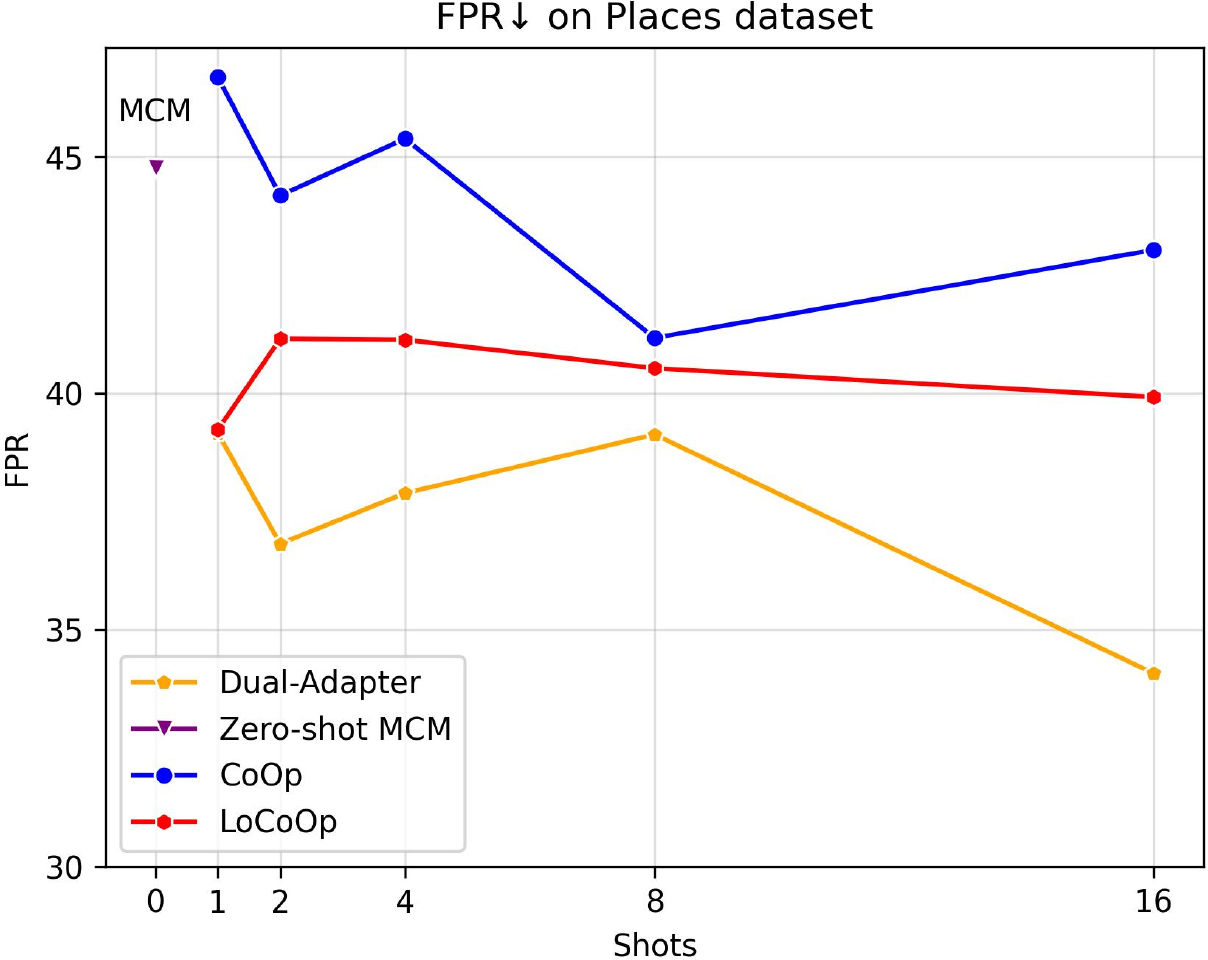}
        \label{fig:results-6}
        \caption{Places FPR}
    \end{subfigure}
    \begin{subfigure}{0.245\linewidth}
        \captionsetup{aboveskip=-2pt, belowskip=0pt}
        \includegraphics[width=0.99\linewidth]{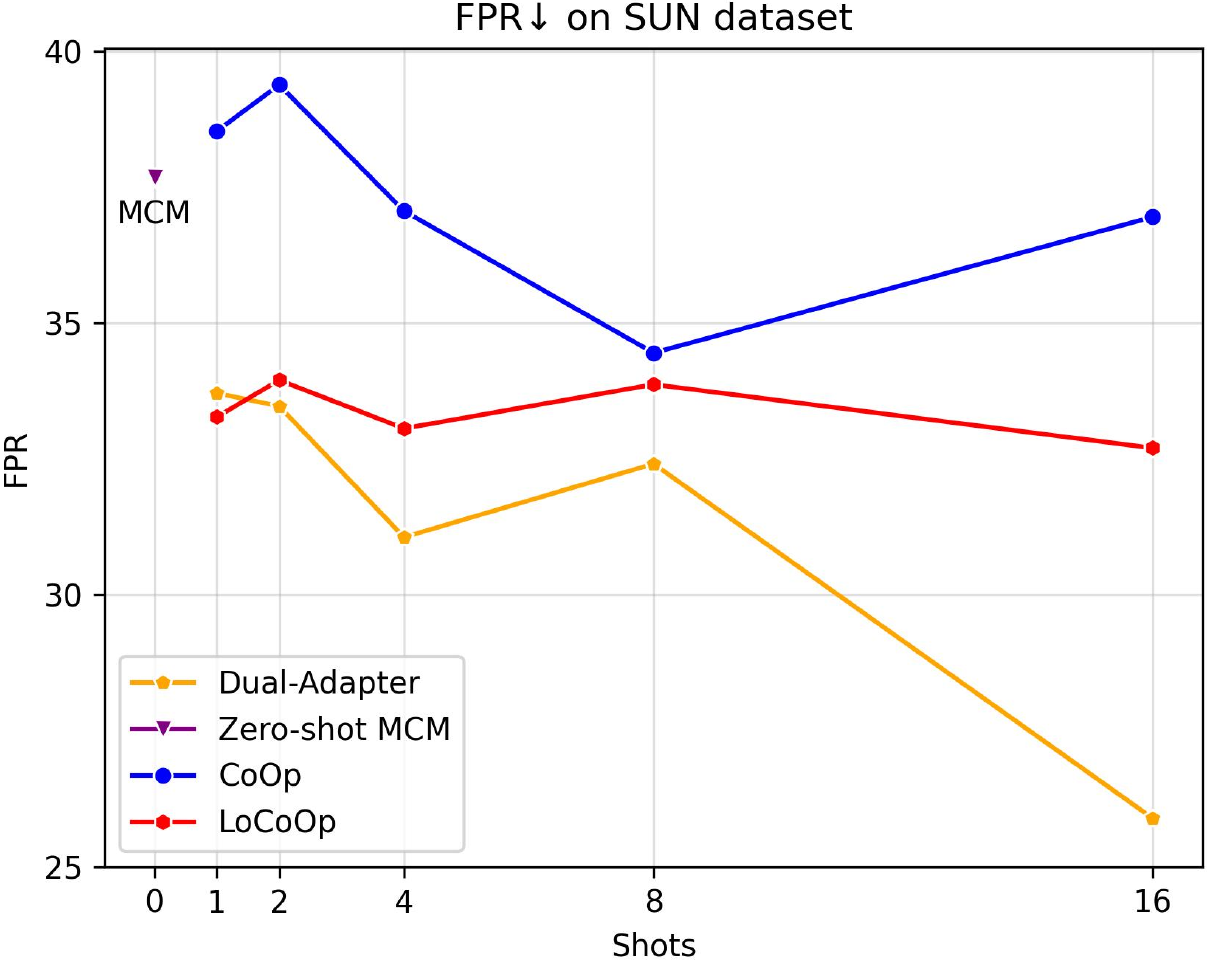}
        \label{fig:results-7}
        \caption{SUN FPR}
    \end{subfigure}
    \begin{subfigure}{0.245\linewidth}
        \captionsetup{aboveskip=0pt, belowskip=0pt}
        \includegraphics[width=0.99\linewidth]{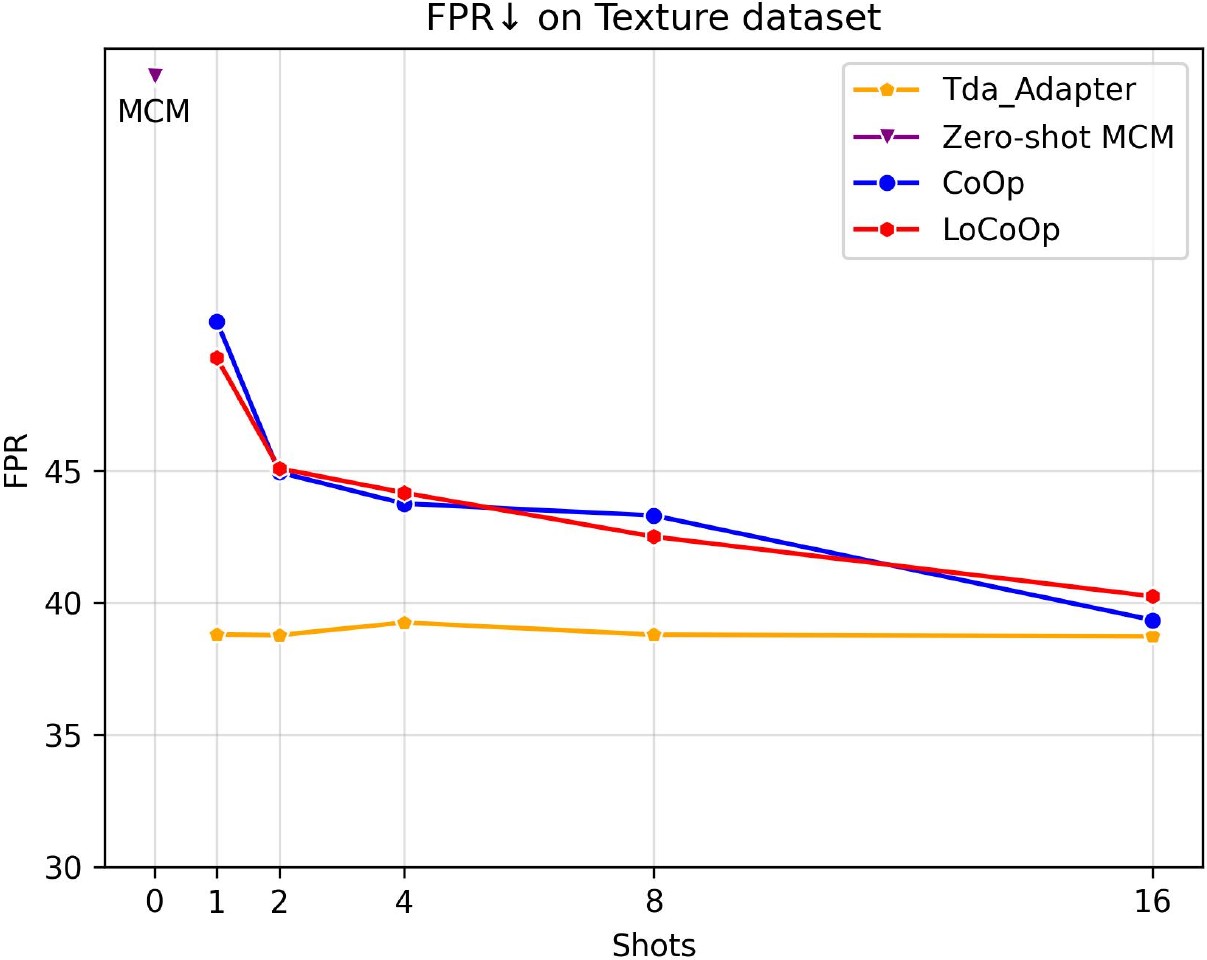}
        \label{fig:results-8}
        \caption{Texture FPR}
    \end{subfigure}
    \caption{\textbf{Few-Shot OOD detection performance of Dual-Adapter and other methods} on the ID dataset  (ImageNet-1k) and four OOD datasets}
    \label{fig:results}
\end{figure}

\subsection{Results}\label{Results}
The results of OOD detection are shown in Fig. \ref{fig:motivation} and Fig. \ref{fig:results}. We compare Dual-Adapter with state-of-art (SOTA) methods. 

In Fig. \ref{figure:1-b} and Fig. \ref{figure:1-d}, we demonstrate that the Dual-Adapter consistently outperforms CoOp, LoCoOp, and MCM in terms of the FPR95 metric across 1, 2, 4, 8, and 16-shot settings.  Furthermore, the same figure illustrates that the Dual-Adapter also exceeds these models in the AUROC metric across 1, 2, 4, and 16-shot settings, while achieving comparable performance in the 8-shot setting. Specifically, the Dual-Adapter recorded an average FPR95 score of 30.50\% and an AUROC of 93.62\%, surpassing LoCoOp by 3.48\% and 0.97\%, respectively. It outperformed CoOp by 8.02\% and 1.69\%, and exceeded MCM by 12.32\% and 2.86\%. 

In Fig. \ref{fig:results}, we present detailed results for the AUROC and FPR metrics across various OOD datasets, including iNatualist \cite{van2018inaturalist}, SUN \cite{xiao2010sun}, Places \cite{zhou2017places}, Texture \cite{cimpoi2014describing}.  Dual-Adapter demonstrates a distinct trend of improving AUROC scores with the addition of more shots, suggesting an enhanced ability to effectively identify out-of-distribution  (OOD) instances as more labeled examples are provided.  The FPR metric remains low and stable across various shot settings, highlighting the model's precision and its capability to effectively reduce false positives. Additionally, detailed experimental data are presented in Table \ref{tab:A}.

% \begin{figure}[t]
%     \centering
%     \includegraphics[width=0.99\linewidth]{img/results.pdf}
%     \caption{\textbf{Few-Shot OOD detection performance of Dual-Adapter and other methods} on the ID dataset  (ImageNet-1k) and four OOD datasets}
%     \label{fig:results}
% \end{figure}

\section{Analysis}

\subsection{Computational Efficiency}
In our study, we conducted a detailed comparison of computational consumption across a variety of machine learning methods. This included zero-shot approaches such as MCM \cite{ming2022delving}, as well as a series of prompt learning methods: CoOp \cite{zhou2022learning}, LoCoOp \cite{miyai2024locoop}, and CoCoOp \cite{zhou2022conditional}. All experiments were performed on a Tesla V100-SXM2-32GB GPU to ensure consistency and reliability in our findings.

\begin{figure}[t]
  \centering
  
  \begin{subfigure}{0.49\linewidth}
    \includegraphics[width=0.99\linewidth]{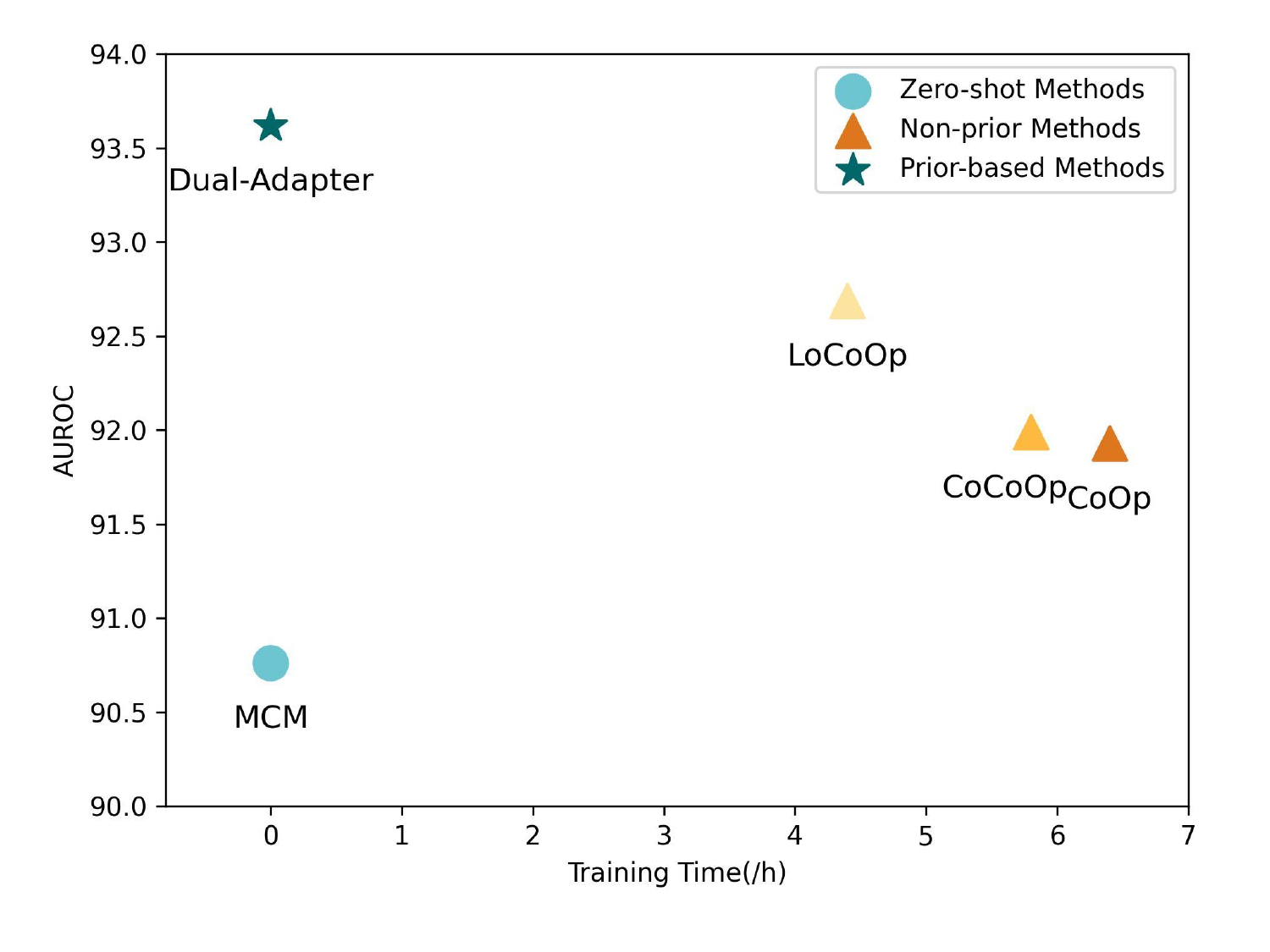}
    \caption{Comparison of AUROC$\uparrow$}
    \label{figure:efficiency-a}
  \end{subfigure}
  \begin{subfigure}{0.49\linewidth}
    \includegraphics[width=0.99\linewidth]{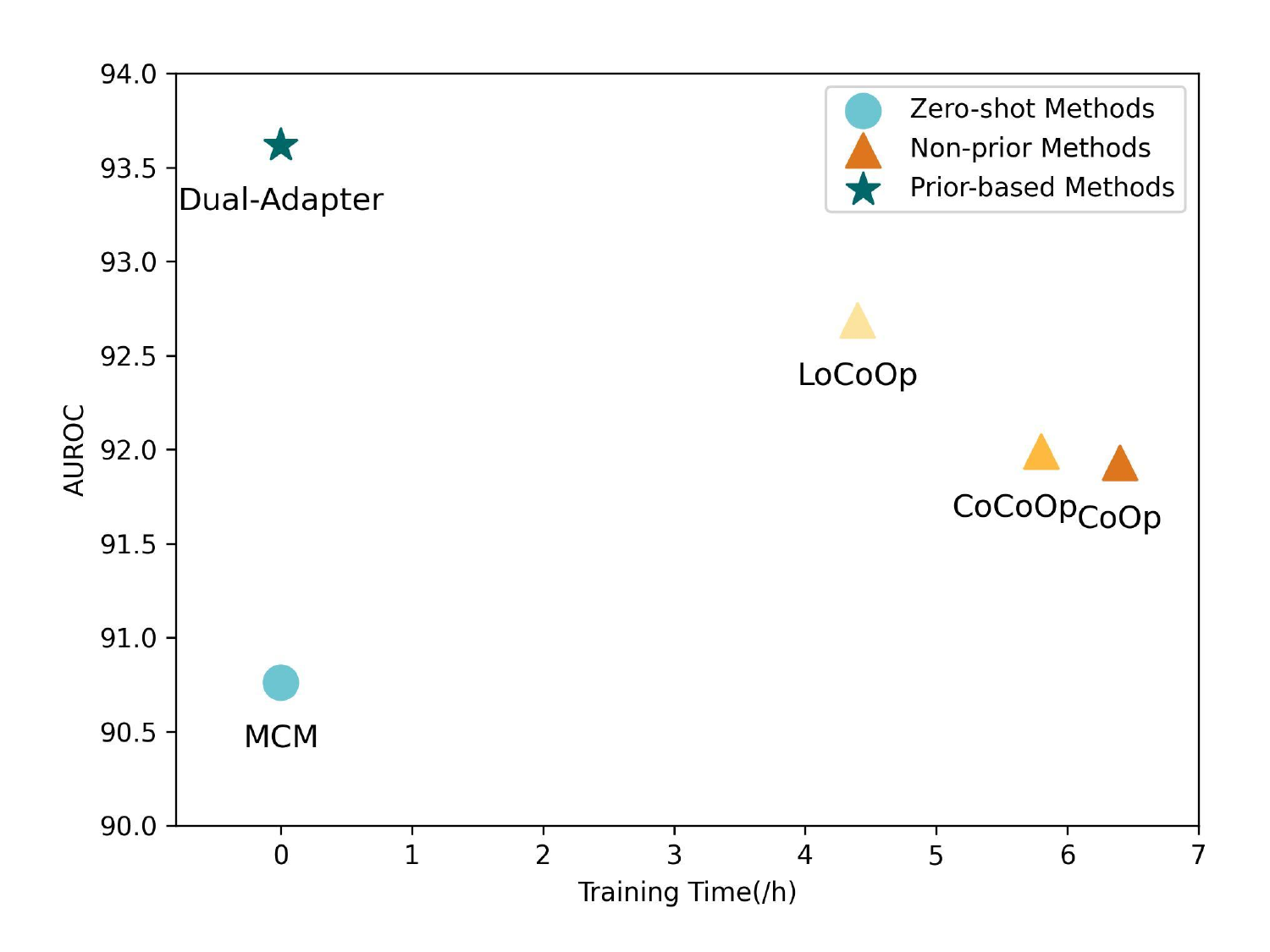}
    \caption{Comparison of FPR95$\downarrow$}
    \label{figure:efficiency-b}
  \end{subfigure}
  
  \caption{Comparative analysis of OOD detection performance and training time on the 16-shot ImageNet \cite{deng2009imagenet} dataset reveals that our Dual-Adapter model achieves superior performance while maintaining high implementation efficiency.}
  \label{figure:efficiency}
\end{figure}

The comparative results are depicted in Fig. \ref{figure:efficiency}. From the findings, it is evident that both the Dual-Adapter and MCM models do not require any training, offering a distinct advantage in scenarios where rapid deployment is necessary. Despite this similarity, the Dual-Adapter model demonstrates a notable superiority over MCM. Dual-Adapter significantly outperforms MCM, achieving 2.86\% higher in AUROC and 12.32\% lower in FPR. 

While the prompt learning methods, CoOp, LoCoOp and CoCoOp, also deliver commendable performance outcomes, they require extensive training times, which could be a limiting factor in time-sensitive applications.  Furthermore, even with this substantial investment in training, they still do not reach the performance levels of the Dual-Adapter model. This shows that the Dual-Adapter model is more efficient, achieving better results faster than the prompt learning models.

\begin{table}[t]
\centering
\caption{Ablation study for Dual-Adapter}
\label{tab:1}
\resizebox{\textwidth}{!}{
\begin{tabular}{lcccccccccc}
\toprule[1pt]
\multirow{2}{*}{\textbf{Method}} & \multicolumn{2}{c}{\textbf{iNaturalist} } & \multicolumn{2}{c}{\textbf{SUN} }   & \multicolumn{2}{c}{\textbf{Places} } & \multicolumn{2}{c}{\textbf{Texture} } & \multicolumn{2}{c}{\textbf{Average} } 

\\ \cmidrule (lr){2-3}  \cmidrule (lr){4-5}  \cmidrule (lr){6-7}  \cmidrule (lr){8-9} \cmidrule (lr){10-11}
  & {FPR95$\downarrow$ } & {AUROC$\uparrow$} & {FPR95$\downarrow$} & {AUROC$\uparrow$} & {FPR95$\downarrow$} & {AUROC$\uparrow$} & {FPR95$\downarrow$} & {AUROC$\uparrow$} & {FPR95$\downarrow$} & {AUROC$\uparrow$} \\ \midrule
\textbf{Positive-Adapter}            & 25.28          & 95.09         & 28.17         & 93.93          & 37.50           & 90.69        & 51.37           & 87.75         & 35.58         & 91.87            \\
\textbf{Negative-Adapter}            & 62.40         & 88.52          & 44.96           & 91.09           & 47.17          & 89.00         & 45.89          & 88.99           & 50.11           & 89.40           \\
\textbf{Dual-Adapter}  & \textbf{23.33}  & \textbf{95.22}  & \textbf{25.89}   & \textbf{94.44}  & \textbf{34.08}   & \textbf{91.55}  & \textbf{38.72}   & \textbf{91.45}  & \textbf{30.50}  & \textbf{93.62}   \\ \bottomrule[1pt]
\end{tabular}}
\end{table}

\begin{table}[t]
\centering
\caption{Comparision results with CLIP-ResNet50. Compared to the zero-sample approach MCM, the mean FPR95 decreased by 7.08\% and the mean AUROC increased by 1.22\%.}
\label{tab:RN50}
\resizebox{\textwidth}{!}{
\begin{tabular}{lcccccccccc}
\toprule[1pt]
\multirow{2}{*}{\textbf{Method}} & \multicolumn{2}{c}{\textbf{iNaturalist} } & \multicolumn{2}{c}{\textbf{SUN} }   & \multicolumn{2}{c}{\textbf{Places} } & \multicolumn{2}{c}{\textbf{Texture} } & \multicolumn{2}{c}{\textbf{Average} } 

\\ \cmidrule (lr){2-3}  \cmidrule (lr){4-5}  \cmidrule (lr){6-7}  \cmidrule (lr){8-9} \cmidrule (lr){10-11}
  & {FPR95$\downarrow$ } & {AUROC$\uparrow$} & {FPR95$\downarrow$} & {AUROC$\uparrow$} & {FPR95$\downarrow$} & {AUROC$\uparrow$} & {FPR95$\downarrow$} & {AUROC$\uparrow$} & {FPR95$\downarrow$} & {AUROC$\uparrow$} \\ \midrule
\multicolumn{11}{c}{Backbone ResNet50} \\
\textbf{MCM \cite{ming2022delving}}  & 31.98  & 93.86  & 46.09  & 90.75  & 60.56  & 85.67     & 60.00   & 85.72  & 49.66  & 89.00 \\
\textbf{Dual-Adapter}  & 40.89  & 91.48  & 38.99  & 91.17  & 47.81   & 87.79  & 42.62   & 90.44  & 42.58  & 90.22   \\ \midrule
\multicolumn{11}{c}{Backbone ViT-B/16} \\
\textbf{MCM \cite{ming2022delving}} & 30.94  & 94.61 & 37.67 & 92.56 & 44.76 & 89.76  & 57.91           & 86.10  & 42.82 & 90.76 \\
\textbf{Dual-Adapter} & \textbf{23.33} & \textbf{95.22} & \textbf{25.89} & \textbf{94.44} & \textbf{34.08} & \textbf{91.55} & \textbf{38.72} & \textbf{91.45} & \textbf{30.50} & \textbf{93.62} \\ 
\bottomrule[1pt]
\end{tabular}}
\end{table}

% Fine-tuning and inferring time
% 跑LoCoOp的训练时长和推理时长

\subsection{Ablation Study}
% pos, neg, pos+neg

\paragraph{The effectiveness of Positive-Adapter and Negative-Adapter}

In this section, we examine a study of the impacts of Positive-Adapter and Negative-Adapter whthin our framework. We conduct experiments using Positive-Adapter, Negative-Adapter, and their combined form, Dual-Adapter, in a 16-shot ImageNet setting. The results are shown in Table \ref{tab:1}. Positive-Adapter is specifically designed to handle positive features, achieving high performance while still offering opportunities for further enhancement. Conversely, Negative-Adapter is tailored to manage negative features but exhibits lower performance in comparison. However, the Negative-Adapter serves as a valuable complement to the Positive-Adapter. Dual-Adapter exhibits superior performance compared to the Positive-Adapter and Negative-Adapter. Both Positive-Adapter and Negative-Adapter have distinct effects, and Dual-Adapter cannot function without either.

\paragraph{Backbone structure} 
% 换成RN50 16-shot的跑数据
We also investigate the effectiveness of Dual-Adapter with CNN architectures.  While the primary experiments utilize a Vision Transformer architecture as the backbone, in line with previous research  \cite{miyai2024locoop}. Many researches based on CNN architecture have also made bright results, so we did supplementary experiments using the RN50 architecture. As Table \ref{tab:RN50} shown, compared to zero-shot MCM \cite{ming2022delving} method, Dual-Adapter achieves 1.22\% improvement in average AUROC and 7.08\% reduction in average FPR on 16-shot setting. This demonstrates that Dual-Adapter performs well under different network backbones.

\subsection{Visualization of Positive Channels and Negative Channels}

\begin{figure}[t]
    \centering
    \begin{subfigure}{1\linewidth}
        \captionsetup{aboveskip=0pt, belowskip=0pt}
        \includegraphics[width=0.99\linewidth]{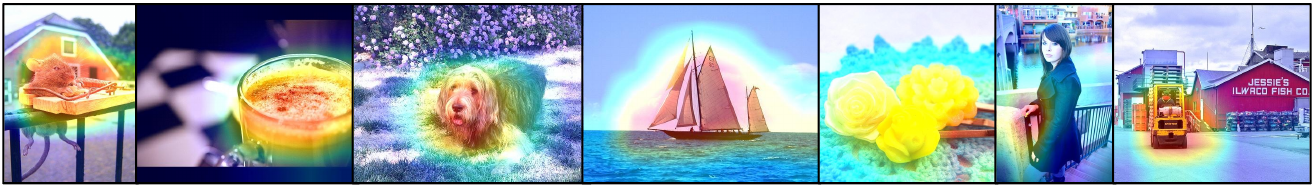}
        \label{fig:channels_positive}
        \caption{Extracted positive channels}
    \end{subfigure}
    \begin{subfigure}{1\linewidth}
        \captionsetup{aboveskip=0pt, belowskip=0pt}
        \includegraphics[width=0.99\linewidth]{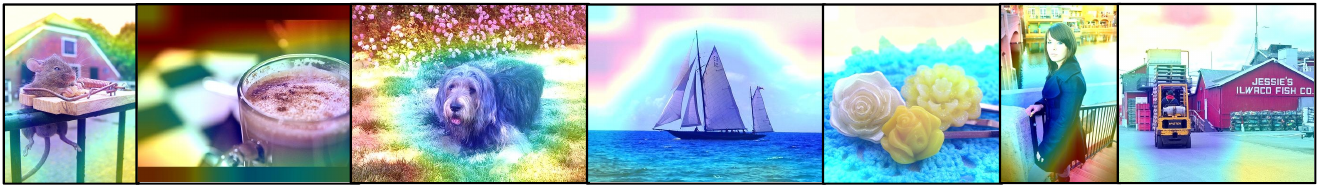}
        \label{fig:channels_negative}
        \caption{Extracted negative channels}
    \end{subfigure}
    \caption{Heatmap for extracted channels. Note that the above images are only a partial listing of positive and negative channels.}
    \label{fig:channels}
\end{figure}

In Fig. \ref{fig:channels}, we categorize the original channels into two groups: positive channels and negative channels. We then visualize each group by displaying heatmaps that provide a visual representation of the data. These heatmaps show that the positive channels accurately represent the main subject in the image, while the negative channel identify objects that are not part of the main focus or are in the background. This validates the underlying logic of the dual-adapter to some extent.

\subsection{Limitation and Outlook} \label{limitation and outlook}

Despite the remarkable results achieved by Dual-Adapter, there are still limitations that need to be addressed in the future: 
 (1) The current approach relies heavily on pre-training knowledge from CLIP. When pre-trained model falis, Dual-Adapter may then fail. To enhance performance further, future research should explore integrating more robust foundational models. Alternatively, Dual-Adapter should its reliance on pre-trained models, so that it's able to cope even if the pre-trained models do not perform well.
 (2) In this study, we primarily focus on classification tasks, but extending our approach to other visual recognition tasks remains an important future direction, such as open world segmentation\cite{cen2021deep}.
 % few-shot ood for 其他视觉任务，引用一下

\section{Conclusion}

% 修改，首次引入， dual cache modeling， 实验
For few-shot out-of-distribution (OOD) detection tasks, we first introduce CLIP prior-based method and propose a simple but effective approach named Dual-Adapter. We introduce the concept of dual cache modeling, segregating features into positive and negative categories to construct caches. This approach leverages previously neglected interfering features to enhance performance. The comprehensive experimental results across benchmark datasets highlight its superiority in enhancing detection performance without the need for additional training. This approach opens new avenues for more robust and efficient OOD detection in practical applications where labeled data is scarce.

\bibliographystyle{unsrtnat}
\bibliography{reference}
% \bibliographystyle{plain}
% \printbibliography
% %%%%%%%%%%%%%%%%%%%%%%%%%%%%%%%%%%%%%%%%%%%%%%%%%%%%%%%%%%%%

\newpage
\appendix
% \paragraph{{\rm {\bf Appendix}}}

% \renewcommand{\thetable}{\Alph{table}}

\section{Detailed Results on few-shot OOD detection}

In this section, we present detailed and accurate results for out-of-distribution  (OOD) detection across various shot settings, including 1, 2, 4, 8, and 16 shots. These results are compared with those from other methods such as MCM, CoOp, and LoCoOp.

\begin{table}[htbp]
\centering
\caption{Few-shot OOD detection with different numbers of ID samples.}
\label{tab:A}
\resizebox{\textwidth}{!}{
\begin{tabular}{lcccccccccc}
\toprule[1pt]
\multirow{2}{*}{\textbf{Method}} & \multicolumn{2}{c}{\textbf{iNaturalist} } & \multicolumn{2}{c}{\textbf{SUN} }   & \multicolumn{2}{c}{\textbf{Places} } & \multicolumn{2}{c}{\textbf{Texture} } & \multicolumn{2}{c}{\textbf{Average} } 

\\ \cmidrule (lr){2-3}  \cmidrule (lr){4-5}  \cmidrule (lr){6-7}  \cmidrule (lr){8-9} \cmidrule (lr){10-11}
  & {FPR95$\downarrow$ } & {AUROC$\uparrow$} & {FPR95$\downarrow$} & {AUROC$\uparrow$} & {FPR95$\downarrow$} & {AUROC$\uparrow$} & {FPR95$\downarrow$} & {AUROC$\uparrow$} & {FPR95$\downarrow$} & {AUROC$\uparrow$} \\ \midrule
    \multicolumn{11}{c}{\textit{zero-shot}} \\
    \textbf{MCM} & 30.94 & 94.61 & 37.67 & 92.56 & 44.76 & 89.76 & 57.91 & 86.10 & 42.82 & 90.76 \\ \midrule

  \multicolumn{11}{c}{\textit{1-shot (one label per class)}} \\ 
\textbf{CoOp}            & 43.38           & 91.26         & 38.53          & 91.95           & 46.68            & 89.09          & 50.64           & 87.83   & 44.81  &90.03            \\
\textbf{LoCoOp}            & 38.49           & 92.49          & 33.27          & 93.67           & 39.23           & 91.07          & 49.25           &  89.13          & 40.17           & 91.53            \\
\textbf{Dual-Adapter}     & 26.26  & 94.64  & 33.71   & 92.92  & 39.16   & 90.59  & 38.79   & 90.98  & 34.38  & 92.28  \\ \midrule

 \multicolumn{11}{c}{\textit{2-shot (two label per class)}} \\ 
\textbf{CoOp}            & 38.89          & 92.12          & 39.38          & 91.58          & 44.17           & 88.98         & 44.92          & 89.16          & 41.85       & 90.46           \\
\textbf{LoCoOp}            & 35.38          & 92.76         & 33.95         & 93.31          & 41.15          & 90.38         & 45.07           & 89.76    & 38.89           & 91.55           \\
\textbf{Dual-Adapter}    & 26.26 & 94.64 & 33.47 & 92.92 & 36.82 & 90.74 & 38.76 & 91.00 & 33.83 & 92.32   \\ \midrule

  \multicolumn{11}{c}{\textit{4-shot (four label per class)}} \\ 
\textbf{CoOp}  & 35.36  & 92.60  & 37.06 & 92.27 & 45.38 & 89.15 & 43.74 & 89.68  & 40.38 & 90.92  \\
\textbf{LoCoOp} & 29.45 & 93.93 & 33.06  & 93.24  & 41.13   & 90.32  & 44.15   & 90.54  & 36.95  & 92.01  \\
\textbf{Dual-Adapter}   &23.44 & 94.97 & 31.06 & 92.96 & 37.89 & 90.74 & 39.24 & 91.00 & 32.90 & 92.42 \\ \midrule

  \multicolumn{11}{c}{\textit{8-shot (eight label per class)}} \\ 
\textbf{CoOp}    & 35.17  & 92.96 &  34.45 & 92.50  & 41.17 & 89.76  & 43.29  & 89.92  & 38.52  & 91.29   \\
\textbf{LoCoOp}  & 27.12  & 94.60 & 33.87 & 93.23   & 40.53 & 90.53 & 42.49 & 90.98 & 36.00 & 92.34 \\
\textbf{Dual-Adapter} & 26.23 & 94.63 & 32.41 & 92.91 & 39.13 & 90.59 & 38.78 & 90.98 & 34.14 & 92.27 \\ \midrule

  \multicolumn{11}{c}{\textit{16-shot (sixteen label per class)}} \\ 
\textbf{CoOp}   & 28.00 & 94.43 & 36.95 & 92.29 & 43.03 & 89.74 & 39.33 & 91.24 & 36.93 & 91.93   \\
\textbf{LoCoOp}  & \textbf{23.06} & \textbf{95.45} & 32.70 & 93.35 & 39.92 & 90.64 & 40.23 & 91.32 & 33.98 & 92.69  \\
\textbf{Dual-Adapter} & 23.33 & 95.22 & \textbf{25.89} & \textbf{94.44} & \textbf{34.08} & \textbf{91.55} & \textbf{38.72} & \textbf{91.45} & \textbf{30.50} & \textbf{93.62} \\ 
\bottomrule[1pt]
\end{tabular}}
\end{table}

\end{document}